\documentclass[letterpaper, 10 pt, journal, twoside]{IEEEtran}

\usepackage{times,multirow,caption,float}
\usepackage{hyperref} 
\usepackage{wrapfig}
\usepackage{graphicx}
\usepackage{epsfig,xspace,layout}
\usepackage{subfigure}
\usepackage{color}
\usepackage{amsfonts}
\usepackage{times}
\usepackage{amssymb}
\usepackage{amsmath}
\usepackage{rotating}
\usepackage{amsthm}
\usepackage{graphicx}
\usepackage{epsfig}
\usepackage{mathrsfs}
\usepackage{caption}
\usepackage{subfigure}
\usepackage{mathtools}
\usepackage{makeidx}
\usepackage{multirow} 
\usepackage{dblfloatfix}
\usepackage{threeparttable}
\usepackage{dsfont}
\usepackage[font={small}]{caption}
\usepackage{cleveref}
\usepackage{algorithm}
\usepackage{algorithmic}
\usepackage{tikz}
\usetikzlibrary{shapes,arrows,positioning,calc}
\usepackage{siunitx}
\usepackage{booktabs}

\theoremstyle{definition}

\theoremstyle{remark}
\newtheorem{remark}{Remark}

\DeclareMathAlphabet{\mathpzc}{OT1}{pzc}{m}{it}

\DeclareFontFamily{U}{jkpmia}{}
\DeclareFontShape{U}{jkpmia}{m}{it}{<->s*jkpmia}{}
\DeclareFontShape{U}{jkpmia}{bx}{it}{<->s*jkpbmia}{}
\DeclareMathAlphabet{\mathfrak}{U}{jkpmia}{m}{it}

\title{CTS: 
Concurrent Teacher-Student Reinforcement Learning for Legged Locomotion}

\author{Hongxi Wang$^{1,*}$, Haoxiang Luo$^{1,*}$, Wei Zhang$^{1,3}$ and Hua Chen$^{2,3}$
\thanks{Manuscript received: May, 14, 2024; Revised July, 27, 2024; Accepted August, 20, 2024.}
\thanks{This paper was recommended for publication by Editor Abderrahmane Kheddar upon evaluation of the Associate Editor and Reviewers' comments. This work was supported in part by the National Natural Science Foundation of China (Grant No. 62073159, and Grant No. 62003155), and in part by the Shenzhen Key Laboratory of Control Theory and Intelligent Systems, under Grant No. ZDSYS20220330161800001. (\textit{Corresponding Author: Hua Chen})} 
\thanks{$^*$: these authors contributed equally.}
\thanks{$^1$ School of System Design and Intelligent Manufacturing (SDIM), Southern University of Science and Technology, Shenzhen, China. Emails: {\tt\footnotesize \{12332640, 12232312\}@mail.sustech.edu.cn, zhangw3@sustech.edu.cn}}
\thanks{$^2$ Zhejiang University-University of Illinois Urbana-Champaign Institute (ZJUI), Haining, China. Email: {\tt\footnotesize huachen@intl.zju.edu.cn}}
\thanks{$^3$ LimX Dynamics, Shenzhen, China.}
\thanks{Digital Object Identifier (DOI): see top of this page.}
}

\usepackage{algorithm}
\usepackage{algorithmic}
\usepackage{graphicx}
\usepackage{subfigure}
\usepackage{cite}

\begin{document}

\maketitle

\begin{abstract}
Thanks to recent explosive developments of data-driven learning methodologies, reinforcement learning (RL) emerges as a promising solution to address the legged locomotion problem in robotics. In this paper, we propose CTS, a novel Concurrent Teacher-Student reinforcement learning architecture for legged locomotion over uneven terrains. Different from conventional teacher-student architecture that trains the teacher policy via RL first and then transfers the knowledge to the student policy through supervised learning, our proposed architecture trains teacher and student policy networks concurrently under the reinforcement learning paradigm. To this end, we develop a new training scheme based on a modified proximal policy gradient (PPO) method that exploits data samples collected from the interactions between both the teacher and the student policies with the environment. 
The effectiveness of the proposed architecture and the new training scheme is demonstrated through substantial quantitative simulation comparisons with the state-of-the-art approaches and extensive indoor and outdoor experiments with quadrupedal and point-foot bipedal robot platforms, showcasing robust and agile locomotion capability. 
Quantitative simulation comparisons show that our approach reduces the average velocity tracking error by up to 20\% compared to the two-stage teacher-student, demonstrating significant superiority in addressing blind locomotion tasks. Videos are available at \textcolor{blue}{\href{https://clearlab-sustech.github.io/concurrentTS/}{https://clearlab-sustech.github.io/concurrentTS/}}.
\end{abstract}


\section{Introduction}\label{sec:introduction}
\IEEEPARstart{L}{comotion} is one of the most important skills for legged robots, which enables them to traverse complicated terrains to accomplish various tasks~\cite{ANYmal,ANYmalField,HOUND,miniCheetah3,miniCheetah,gong2019feedback,hong2022three}. Due to the complicated contact interaction with the uneven terrain and the inherent nonlinear and hybrid dynamics, the locomotion controller synthesis problem is widely acknowledged to be challenging~\cite{wensing2022optimizationbased}. Recently, reinforcement learning-based methods have been shown to be a promising solution to legged locomotion and have achieved remarkable results ~\cite{joonholee2019scirobotics,siekmann2020learning,siekmann2021blind, kumar2021rma, margolisyang2022rapid,Hwangbo2022vel_est, rudin2022learn_in_minutes}.

\begin{figure}[t]
    \centering
    \includegraphics[width=1.0\linewidth]
    {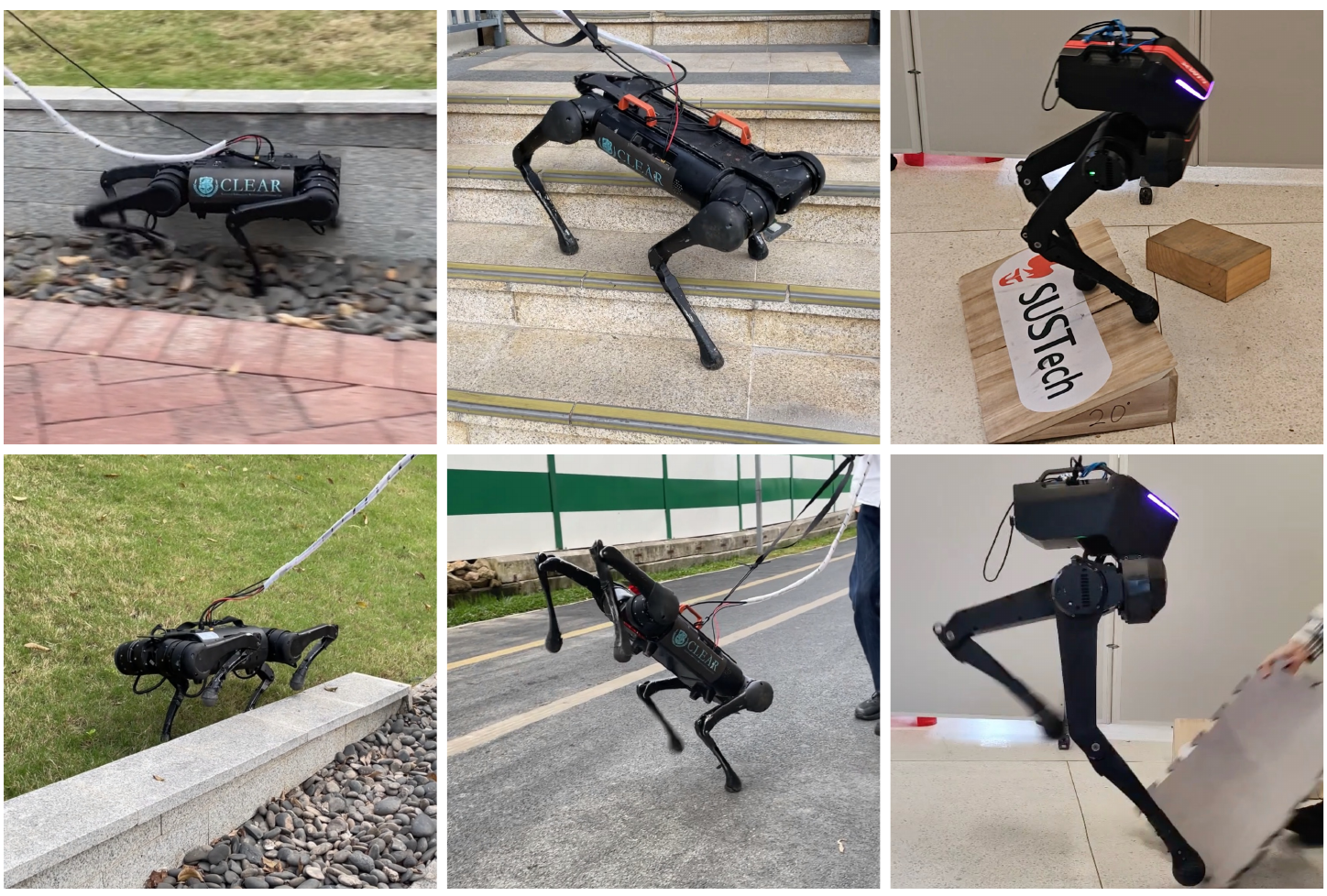} 
    \caption{CTS enables legged robots of various sizes and configurations to achieve robust and agile locomotion across challenging real-world terrains, while also possessing exceptional capabilities to withstand strong external disturbances.}
    \label{fig:real_world}
    \vspace{-15px}
\end{figure}



\begin{figure*}[b]
    \centering
    \includegraphics[width=0.8\linewidth]{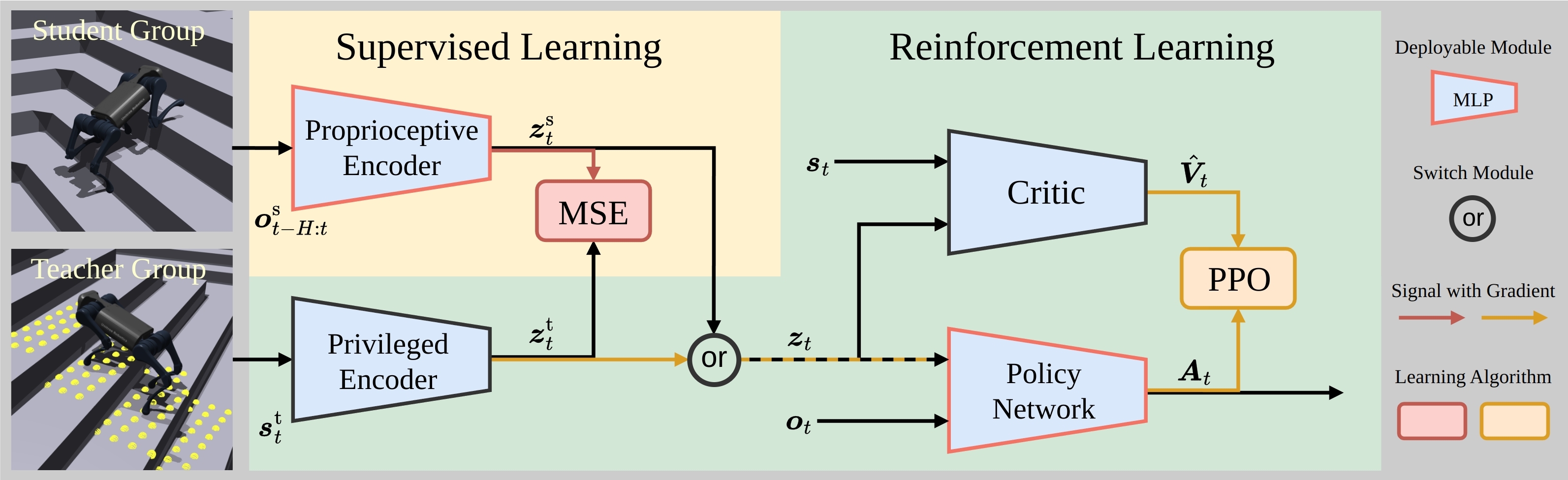} 
    \caption{\footnotesize Overview of the learning framework. The teacher and student policies are trained concurrently using PPO within an asymmetric actor-critic framework. Agents in both groups share the same critic and policy network, with actions determined by observations and latent representations from either privileged or proprioceptive encoder. The privileged encoder is trained via policy gradient, while the proprioceptive encoder undergoes supervised learning to minimize reconstruction loss.}
    \label{fig:framework}
    \vspace{-15px}
\end{figure*}
Nowadays, the teacher-student paradigm is one of the most widely adopted and studied learning-based methodologies for achieving legged locomotion~\cite{joonholee2020scirobotics, kumar2021rma, wu2023ts_amp}. In such a paradigm, a teacher policy having full access to all locomotion-related (\emph{privileged}) information (e.g., terrain details, contact information, accurate inertial parameters) is first trained by reinforcement learning. Then, a student policy that operates purely with proprioceptive feedback is trained by supervised learning to reconstruct the latent representation from the encoder of trained teacher policy and/or imitate the actor output of teacher policy. This approach allows for efficient learning and sim-to-real transfer where the robot operates with only proprioception, enabling various legged robots to locomote in complex terrains~\cite{kumar2021rma,wu2023ts_amp,zjuBipedal}. 
\cite{wu2023ts_amp} integrates teacher-student learning with Adversarial Motion Priors (AMP) to enable quadruped robots to learn natural and robust locomotion. ~\cite{zjuBipedal} leverages the advances in rapid motor adaptation for quadruped locomotion originally proposed in~\cite{kumar2021rma}, and extends the application to humanoid robots. Similar training paradigm has also been shown capable of addressing other challenging locomotion tasks with richer inputs including exteroceptive measurements~\cite{joonholee2022scirobotics,egocentricLoco,anymalParkour,robotParkour,extremeParkour}. In addition to the above two-stage training process, the Regularized Online Adaptation (ROA) method, as proposed in~\cite{roa} integrates the training of encoders of the teacher and student policies into a single stage, achieving mobile-manipulation tasks. During the iterations in ROA where the policy uses the output of the proprioceptive encoder as input, the policy network itself is not updated. In these iterations, only the proprioceptive encoder undergoes supervised learning. This means the policy network does not undergo reinforcement learning training conditioned on the proprioceptive encoder's input.

Different from the teacher-student paradigm, representation learning leverages the idea of dynamics learning to enhance legged locomotion~\cite{nahrendra2023dreamwaq, long2024hybrid}. DreamWaQ~\cite{nahrendra2023dreamwaq} exploits learned representations through variational autoencoders (VAE) to improve legged locomotion performance. Hybrid Internal Model~\cite{long2024hybrid} (HIM) treats external states such as terrain conditions as disturbances and learns representations through contrastive learning.
In this way, the HIM method closes the gap between the representations generated from historical sequence of observations and future observations, aiding in obtaining latent representations that include system dynamics to assist in reinforcement learning training. A common characteristic of these methodologies is the use of a regression objective for state representation, which requires the neural network to match the targets accurately.


In this work, we explore the mechanism for systematically integrating the information from the teacher strategy with that from the student strategy. To this end, we propose a concurrent teacher-student learning architecture for legged locomotion. Such an architecture philosophically combines the teacher-student paradigm with the core of representation learning and embodies the training of teacher and student policies concurrently, making it more streamlined than two-stage training. More importantly, we guide the student's training using reinforcement learning objectives instead of merely imitating the teacher. Since the differences in observations between the student and the teacher make perfect imitation difficult, the goal of purely imitating the teacher may not be optimal. Incorporating reinforcement learning objectives during the student's training helps in deriving a better policy. We validate our approach on multiple hardware platforms including quadrupeds of different sizes and a more challenging, underactuated bipedal robot with point feet. Furthermore, robots have demonstrated the ability to navigate through challenging terrains and against external disturbances.

The contributions of this work are summarized as follows. First, the proposed CTS reinforcement learning architecture effectively exploits the interplay between teacher and student networks to enhance the overall performance of the resultant policy. Simulation experiments quantitatively show that the proposed CTS architecture achieves an improvement of up to 20\% in velocity tracking error on uneven terrains compared to state-of-the-art teacher-student and ROA methods. 
Second, we conducted extensive real-world experiments with various hardware platforms to demonstrate the small sim-to-real gap of the proposed architecture. The extensive hardware experimental results with various hardware platforms, including quadrupeds and point-foot bipeds, illustrate the capability of our proposed method to enable robust locomotion over challenging terrains in both indoor and outdoor environments.


\section{Concurrent Teacher-Student Reinforcement Learning for Legged Locomotion}\label{sec:method}

An overview of the proposed concurrent teacher-student reinforcement learning architecture for legged locomotion is shown in Fig.~\ref{fig:framework}.
In this section, we formulate our problem and develop the proposed architecture. 

\subsection{Legged Locomotion Problem and Reinforcement Learning}
In essence, legged locomotion focuses on finding the appropriate joint torque commands for all actuated joints of the robots given the sensory measurements. Under the assumption of only having accessibility to proprioceptive measurements from IMU and joint encoders, 
the legged locomotion dynamics can be formulated as an following infinite-horizon partially observable Markov decision process (POMDP), defined by the tuple $\left<\mathcal{S},\mathcal{A},\mathcal{O},T,\Omega,R\right>$, where $\mathcal{S} \subset \mathbb{R}^n$ is the set of full state including all dynamic information of legged robot and environment around, $\mathcal{A} \subset \mathbb{R}^m$ is the set of action, $\mathcal{O} \subset \mathbb{R}^o$ is the set of observation, $T(\boldsymbol{s}',\boldsymbol{s},\boldsymbol{a})=p(\boldsymbol{s}'|\boldsymbol{s},\boldsymbol{a})$ is the state transition function, $\Omega(\boldsymbol{o},\boldsymbol{s},\boldsymbol{a})=p(\boldsymbol{o}|\boldsymbol{s},\boldsymbol{a})$ is the observation function and $R(\boldsymbol{s},\boldsymbol{a},\boldsymbol{s}')$ is the reward function. Our goal is to find the optimal policy $\pi^*$ to maximize the expected discounted return over the trajectory:
\begin{equation}
    J(\pi)=\mathbb{E}_{\pi}\left[\sum_{t=0}^\infty \gamma^t R(\boldsymbol{s},\boldsymbol{a},\boldsymbol{s}')\right]
\end{equation}
with discounting factor $\gamma \in [0,1]$.

\textbf{State Space:} We denote by $\boldsymbol{o}_t$ the observation and $\boldsymbol{s}_t$ the state at time $t$. The proprioceptive observation $\boldsymbol{o}_t \in \mathbb{R}^n$ consists of angular velocity, gravity vector in the base frame of the robot, joint positions and velocities, command, and previous actions. Here, $n$ denotes the dimensionality of the proprioceptive observation vector, encompassing all aforementioned components. The command consists of the desired velocity $v^\text{cmd}_x,v^\text{cmd}_y$ and the angular velocity $\omega^\text{cmd}_z$ in the base frame. The full state $\boldsymbol{s}_t$ consists of proprioceptive observation $\boldsymbol{o}_t$, base linear velocity $\boldsymbol{v}_t \in \mathbb{R}^3$, terrain height samples $i_t \in \mathbb{R}^m$ and other privileged information including feet contact forces, joint torques and acceleration of joints. In this context, $m$ represents the number of terrain height samples collected, forming a vector that describes the terrain profile.

\textbf{Action Space:} For each actuated joint, the action $\boldsymbol{a}_t \in \mathbb{R}^k$ represents the angular deviation of the robot's joint relative to its nominal position , where $k$ denotes the number of actuated joints. Hence, the robot's joint PD controller reference is
\begin{equation}
    \boldsymbol{q}_t^\text{ref} = \boldsymbol{q}^\text{nominal} + K\boldsymbol{a}_t
\end{equation}
with some scale $K$.

\subsection{Concurrent Teacher-Student Architectures}\label{sec:architectures}
To learn the optimal policy, the robot needs to infer its current state $\boldsymbol{s}_t$ from the available observation $\boldsymbol{o}_t$. It is in general impossible to infer the actual state from a single observation, due to the partial observability of the environment. Thus, the inference problem $p(\boldsymbol{s}_t|\boldsymbol{o}_t,\boldsymbol{o}_{t-1},\cdots,\boldsymbol{o}_{t-n})$ requires the historical sequence of observation. Recent works have leveraged variational autoencoder (VAE) or teacher-student learning to implicitly infer state- or task-relevant information. We integrated the advantages of both approaches and employed Proximal Policy Optimization (PPO) for training.

We train the teacher and student policies concurrently by dividing parallel agents into two groups named teacher group and student group, then employing the asymmetric actor-critic framework shown in Fig.~\ref{fig:framework}, where the teacher policy includes privileged encoder and policy network, and the student policy includes proprioceptive encoder and policy network.
Agents in two groups are both trained using proximal policy optimization (PPO) while they share the same policy network $\pi_\theta$ and critic network $V_\phi$. The policy network outputs action $\boldsymbol{a}_t$ given proprioceptive observation $\boldsymbol{o}_t$ and latent representation $\boldsymbol{z}_t \in \mathbb{R}^{32}$, which were generated by the precedent encoder. The latent representation $\boldsymbol{z}_t$ is mapped onto a unit hypersphere with a $\mathcal{L}_2$-normalization. The critic network estimates the state value for some given state $\boldsymbol{s}_t$ and latent representation $\boldsymbol{z}_t$.

Throughout this manuscript, we use superscripts $(\cdot)^\text{t}$ and $ (\cdot)^\text{s}$ to indicate teacher group and student group, respectively. Agents in teacher group have access to full state $\boldsymbol{s}_t$ while they utilize privileged encoder $E^{\text{t}}_{\theta}$ to encode state $\boldsymbol{s}^\text{t}_t$ into latent representation $\boldsymbol{z}^\text{t}_t$. Agents in the student group only have access to proprioceptive observation, allowing policies to be deployed in real environments. Proprioceptive encoder $E^{\text{s}}_{\theta}$ is leveraged to encode observation sequence $\boldsymbol{o}^\text{s}_{t-H:t} = [\boldsymbol{o}^\text{s}_t, \cdots, \boldsymbol{o}^\text{s}_{t-H}]^T$ into latent representation $\boldsymbol{z}^\text{s}_t$ similar to privileged encoder. 

The latent representation $\boldsymbol{z}_t$ is generated from the privileged encoder or the proprioceptive encoder, which guides the policy network to produce specific actions for various terrains and scenarios. The privileged encoder and policy network is trained through policy gradient to maximize the expected discounted return, while the proprioceptive encoder is trained through supervised learning to minimize the reconstruction loss between outputs of proprioceptive encoder and privileged encoder. All modules are designed as the Multi-Layer Perceptron (MLP) with Exponential Linear Unit (ELU) activation. More details on each network are shown in Table~\ref{tab:network}.
{\setlength{\tabcolsep}{5pt}
\begin{table}[b]
    \centering
    \caption{Network Architectures}
    \label{tab:network}
    \begin{tabular}{lcccc}
    \toprule[1.1pt]
    Module & Notation & Inputs & Hidden Layer &  Outputs \\ 
    \hline 
    Privileged Enc.  & $E^{\text{t}}_{\theta}$  & $\boldsymbol{s}^\text{t}_t$ & [512, 256] & $\boldsymbol{z}^\text{t}_t$ 
\\
    Proprioceptive Enc.  & $E^{\text{s}}_{\theta}$  & $\boldsymbol{o}^\text{s}_{t-H:t}$  & [512, 256] &  $\boldsymbol{z}^\text{s}_t$\\
    Policy Network  & $\pi_\theta$  & $\boldsymbol{o}_t, \boldsymbol{z}_t$  & [512, 256, 128] &  $\boldsymbol{a}_t$\\
    Critic  & $V_\phi$  & $\boldsymbol{s}_t, \boldsymbol{z}_t$  & [512, 256, 128] & $ \hat V_t$\\
    \bottomrule[1.1pt]
    \end{tabular}
\end{table}}
\vspace{-10px}
\subsection{Training Pipeline}\label{sec:training pipeline}
Due to the structural change of the overall architecture and the way we exploit the information from the teacher and student groups, the conventional training process does not apply directly and hence needs to be adjusted. The training process of proposed concurrent teacher-student pipeline is shown in Algorithm~\ref{alg:pipeline}.
Since agents are divided into two groups, the Monte-Carlo approximation of PPO-Clip objective functions of each group $L^\text{t}(\theta, \theta^\text{t}), L^\text{s}(\theta)$ is defined as:

\begin{equation}
\begin{aligned}
    L^\text{ppo,t}(\theta, \theta^\text{t}) = &\frac{1}{|\mathcal{D}^\text{t}|T} \sum_{\tau \in \mathcal{D}^\text{t}} \sum_{t=0}^T  \\
    & \min \left( r^\text{t}_t \hat A^\text{t}_t, \text{clip}(r^\text{t}_t, 1-\epsilon,1+\epsilon)\hat A^\text{t}_t \right) 
\end{aligned}
\end{equation}
\begin{equation}
\begin{aligned}
    L^\text{ppo,s}(\theta) = &\frac{1}{|\mathcal{D}^\text{s}|T} \sum_{\tau \in \mathcal{D}^\text{s}} \sum_{t=0}^T  \\
    & \min \left( r^\text{s}_t \hat A^\text{s}_t, \text{clip}(r^\text{s}_t, 1-\epsilon,1+\epsilon)\hat A^\text{s}_t \right) 
\end{aligned}
\end{equation}
where $\mathcal{D}^\text{t}$ and $\mathcal{D}^\text{s}$ are sets of teacher group and student group trajectories by interacting with environment using $E^{\text{t}}_{\theta_k}, \pi_{\theta_k}$ and $E^{\text{s}}_{\theta_k}, \pi_{\theta_k}$, respectively. $T$ is the length of corresponding trajectory. $r^\text{t}_t, r^\text{s}_t$ are ratio functions of two groups:
\begin{equation}
    r^\text{t}_t(\theta, \theta^\text{t}) = \frac{\pi_\theta\left(\boldsymbol a^\text{t}_t |  \boldsymbol{o}^\text{t}_t, E^{\text{t}}_{\theta}(\boldsymbol{s}^\text{t}_t)\right)}{\pi_{\theta_\text{old}}\left(\boldsymbol a^\text{t}_t |  \boldsymbol{o}^\text{t}_t, E^{\text{t}}_{\theta_\text{old}}(\boldsymbol{s}^\text{t}_t)\right)}
\end{equation}
\begin{equation}
    r^\text{s}_t(\theta) = \frac{\pi_\theta\left(\boldsymbol a^\text{s}_t | \boldsymbol{o}^\text{s}_t, E^{\text{s}}_{\theta}(\boldsymbol{o}_{t-H:t}^{\text{s}})\right)}{\pi_{\theta_\text{old}}\left(\boldsymbol a^\text{s}_t | \boldsymbol{o}^\text{s}_t, E^{\text{s}}_{\theta}(\boldsymbol{o}_{t-H:t}^{\text{s}})\right)}
\end{equation}

Value function $V_\phi$ is trained by regression on mean-square error between $V_\phi(\boldsymbol s_t,\boldsymbol z_t)$ and $\hat R_t$ estimated by Generalized Advantage Estimation (GAE) using trajectories from both groups. The Monte-Carlo approximation of value loss is defined as:
\begin{equation}
    L^\text{value}(\phi) = \frac{1}{|\mathcal{D}|T}\sum_{\tau \in \mathcal{D}} \sum_{t=0}^T    \left( V_\phi(\boldsymbol s_t,\boldsymbol z_t) - \hat R_t \right) ^2
\end{equation}

In order to make proprioceptive encoder of student learn from privileged encoder of teacher, reconstruction loss is introduced to further update the proprioceptive encoder by minimizing the outputs difference between proprioceptive encoder and privileged encoder. Its monte carlo approximation is defined as:
\begin{equation}
    L^\text{rec}(\theta^\text{s}) = \frac{1}{|\mathcal{D}^\text{s}|T} \sum_{\tau \in \mathcal{D}^\text{s}} \sum_{t=0}^T \left\Vert E^{\text{s}}_{\theta}(\boldsymbol{o}_{t-H:t}^{\text{s}}) - E^{\text{t}}_{\theta}(\boldsymbol{s}^\text{t}_t)\right\Vert_2^2
\end{equation}

\begin{remark}\label{rmk:1}
Different from two-stage teacher-student paradigm in which the student policy aims to imitate a well-trained teacher policy, the proposed CTS extensively exploit the interplay between the teacher and the student policies. Such an online iterative training scheme may seem to cause the student to learn a moving target, but this change is gradual, ultimately leading to better student policy performance as shown later in the results section. 
\end{remark}


\begin{algorithm}[h]
\caption{Concurrent Teacher-Student Training}
\label{alg:pipeline}
\begin{algorithmic}[1]
\STATE{Initialize environment and networks}
\FOR{$k = 0,1,...$}
    \STATE{Collect sets of trajectories $\mathcal{D}^\text{t}$ and $\mathcal{D}^\text{s}$ with latest policy}
    \STATE{Compute $\hat R_t$ and $\hat A_t$ using GAE}
    \FOR{epoch $i = 0,1,...$}
        \STATE{Use $\boldsymbol{\theta}$ represent $\theta^\text{t},\theta$ for notational brevity
        \STATE{$\boldsymbol{\theta} \leftarrow \boldsymbol{\theta} + \alpha_\text{ppo}\nabla_{\boldsymbol{\theta}}\left(L_i^\text{ppo,t}(\boldsymbol{\theta}) + L_i^\text{ppo,s}(\theta) \right)$}
        \STATE{$\phi \leftarrow \phi - \alpha_\text{ppo}\nabla_\phi L_i^\text{value}(\phi)$}
        }
    \ENDFOR
    \FOR{epoch $i = 0,1,...$}
        \STATE{$\theta^\text{s} \leftarrow \theta^\text{s} - \alpha_\text{ts} \nabla_{\theta^\text{s}} L_i^\text{rec}(\theta^\text{s})$}
    \ENDFOR
\ENDFOR
\end{algorithmic}
\end{algorithm}
\vspace{-15px}

\section{Implementation Details}\label{sec:implementation}

\subsection{Reward Design}\label{sec:reward_design}
We employed a unified reward structure adaptable to quadruped robots of various sizes and dynamic parameters. For point-foot bipedal robot, we reused all the reward items from the quadruped system, with the addition of a minimal number of necessary reward terms tailored to its specific locomotive traits. Details of the reward function are presented in Table~\ref{tab:reward}.

To enable the agent to produce smooth and graceful motions, we conducted a comparative analysis on the locomotion characteristics of legged robots based on optimal control and reinforcement learning. We discovered the most significant difference lies in the end-effectors trajectories of the swing legs. The end effectors of the swing legs of legged robots based on reinforcement learning tend to move along the shortest trajectory, keeping them close to the ground throughout. This results in motions that are neither visually appealing nor well-suited for unstructured terrains. In contrast, optimal control-based approaches often plan a smooth curve so that the feet takeoff and touchdown vertically. We proposed a feet regulation reward $r^\text{fr}$ to characterize this feature:
\begin{equation}
\begin{aligned}
    r^\text{fr} &=  \sum_\text{feet} \Vert\boldsymbol{v}^\text{foot}_{xy}\Vert_2^2 \exp\left(-\frac{{p}^\text{foot}_z}{0.025h^{\text{des}}}\right)
\end{aligned}
\end{equation}
where $p^\text{foot}_z, \boldsymbol{v}^\text{foot}, h^{\text{des}}$ are foot height, foot velocity and desired body height with respect to the ground, respectively.

\begin{table}[t]
    \centering
    \caption{Reward Terms}
    \label{tab:reward}
    \begin{tabular}{lll}
    \toprule[1.1pt]
    Reward Term & Equation &  Weight \\ 
    \hline 
    Lin. velocity tracking & $\exp\left( -4\Vert\boldsymbol{v}_{xy}^\text{cmd} - \boldsymbol{v}_{xy}\Vert_2^2 \right)$ &  $1.0$ \\
    Ang. velocity tracking  & $\exp\left( -4({\omega}_{z}^\text{cmd} - {\omega}_{z})^2 \right)$ &  $0.5$ \\
    Lin. velocity ($z$)  & ${v}_{z}^2$ &  $-2.0/\textcolor{red}{-0.5}$ \\
    Ang. velocity ($xy$)  & $\Vert\boldsymbol{\omega}_{xy}\Vert_2^2$ &  $-0.05$ \\
    Joint acceleration & $\ddot{\boldsymbol{q}}^2$ &  $-2.5 \times 10^{-7}$ \\
    Joint power  & $|\boldsymbol{\tau}| |\dot{\boldsymbol{q}}|^T$ &  $-2 \times 10^{-5}$ \\
    Joint Torque & $\Vert\boldsymbol{\tau}\Vert_2^2$ &  $-0.0001$ \\
    Base height & $(h^{\text{des}} - h)^2$ &  $-1$ \\
    Action rate & $\Vert\boldsymbol{a}_t - \boldsymbol{a}_{t-1}\Vert_2^2$ &  $-0.01$ \\
    Action smoothness & $\Vert\boldsymbol{a}_t - 2\boldsymbol{a}_{t-1} - \boldsymbol{a}_{t-2}\Vert_2^2$ &  $-0.01$ \\
    Collision  & $ n_\text{collision} $ &  $-1$ \\
    Joint limit & $ n_\text{limitation} $ &  $-2$ \\
    Feet regulation & $ r^\text{fr} $ &  $-0.05$ \\
    \textcolor{red}{Orientation ($xy$)}  & \textcolor{red}{$\Vert\boldsymbol{g}_{xy}\Vert_2$} &  \textcolor{red}{$-5.0$} \\
    \textcolor{red}{Feet distance}  & \textcolor{red}{$r^\text{fd}$} &  \textcolor{red}{$-100$} \\
    \textcolor{red}{Feet contact force} & \textcolor{red}{$r^\text{ff} $} &  \textcolor{red}{$-2.0$} \\
    \textcolor{red}{Feet velocity} & \textcolor{red}{$r^\text{fv}$} &  \textcolor{red}{$-2.0$} \\
    \bottomrule[1.1pt]
    \end{tabular}
    \begin{tablenotes}
        
        \item Black: reward terms used for both biped and quadruped.
        \item \textcolor{red}{Red: biped modified weights and newly added terms.}
    \end{tablenotes}
    \vspace{-10px}
\end{table}

In consideration of the higher degree of underactuation in bipeds compared to quadrupeds, maintaining base stability presents a more significant challenge. Accordingly, we have carefully decreased the penalty associated with the linear velocity in the Z direction and introduced a base orientation reward to penalize the base roll and pitch angles. This adjustment enables the bipedal robot to maintain a level posture across a variety of complex terrains as effectively as possible. Our observations indicate that the biped tends to favor foot placements closer to the center-line to minimize torque on the center of mass. However, this behavior increases the risk of leg collisions. To mitigate this issue, we have incorporated a feet distance penalty function, denoted as $r^\text{fd}$:
\begin{equation}
\begin{aligned}
     r^\text{fd}=\max\left(0, 0.1-\left\Vert{p}^\text{left}_{xy}-{p}^\text{right}_{xy} \right\Vert_2\right)
\end{aligned}
\end{equation}

Due to the lack of sole structures, the point-foot bipedal robot is extremely difficult to maintain static standing balance. Therefore, we refer to the method in~\cite{siekmann2021sim} and add periodic rewards to encourage the generation of a regular periodical gait for self-balancing. The rewards $r^\text{ff}$ and $r^\text{fv}$ penalize the foot contact forces during the stance phase and foot velocities during the swing phase, respectively, allowing the agent to learn specified contact patterns.
\begin{equation}
\begin{aligned}
     r^\text{ff}=\sum_\text{feet}\left[1-C_i^{\mathrm{des}}\left(\phi_i\right)\right]\left[1-\exp \left(-0.04\left\|\boldsymbol{f}^\text{foot,i}\right\|_2\right)\right]
\end{aligned}
\end{equation}
\begin{equation}
\begin{aligned}
     r^\text{fv}=\sum_\text{feet} C_i^{\mathrm{des}}\left(\phi_i\right)\left[1-\exp \left(-4\left\|\boldsymbol{v}^\text{foot,i}_{xy}\right\|_2\right)\right]
\end{aligned}
\end{equation}
where $\boldsymbol{f}^\text{foot,i}$ represents the ground contact force of the corresponding leg, $i\in\left(\text{left, right}\right)$. The function $C_i^{\mathrm{des}}$ computes the desired foot contact state from the gait phase $\phi_i$, following the method described in ~\cite{wu2023multi_gait}.


%


\subsection{Environment Setup}\label{sec:training setup}
We use IsaacGym simulator~\cite{isaacGym} to train 8192 parallel agents on different terrains, To keep the leadership role of the teacher in skill learning, we distribute the agents among the two groups with a ratio of 3:1, i.e.,  teacher group and student group consist of 6144 and 2048 agents, respectively. 
A better expression: It takes around 3000 iterations for the policy to acquire the ability of handling challenging terrains like stairs, which corresponds to around 105 minutes of wall clock time using a Nvidia RTX 4090 for training
The policy's performance will continue improving with further training. During the training process, we set the maximum time duration for each episode to be 20 seconds, corresponding to 1,000 time steps with a control frequency of 50 Hz. All episodes were terminated upon reaching the maximum time duration or upon experiencing a robot fall-over. The joint PD controller parameters are set to be $k_\text{p} = 20.0$, $k_\text{d} = 0.5$ for A1, $k_\text{p} = 40.0$, $k_\text{d} = 1.0$ for Aliengo, $k_\text{p} = 40.0$, $k_\text{d} = 2.5$ for P1. The length of the observation sequence $\boldsymbol{o}^\text{s}_{t-H:t}$ is set to 5 for both quadruped and bipedal policies, following the settings reported in~\cite{nahrendra2023dreamwaq} The algorithm performed an iteration every 24 time steps. Hyper-parameters for training are presented in Table~\ref{tab:hyper parameter}, and adaptation of the learning rate is similar to~\cite{rudin2022learn_in_minutes}. 

\begin{table}[t]
    \centering
    \caption{Hyper Parameters for Training}
    \label{tab:hyper parameter}
    \begin{tabular}{cc}
    \toprule[1.1pt]
    PPO &   \\ 
    \hline 
    Batch size  & $8192 \times 24$ \\
    Mini-batch size  & $8192 \times 6$ \\
    Number of epochs  & $5$ \\
    Clip range  & $0.2$ \\
    Entropy coefficient  & $0.01$ \\
    Discount factor & $0.99$ \\
    GAE discount factor  & $0.95$ \\
    Desired KL-divergence  & $0.01$ \\
    Learning rate & adaptive  \\
    \toprule[1.1pt]
    Proprioceptive Encoder &   \\ 
    \hline 
    Batch size  & $2048 \times 24$ \\
    Mini-batch size  & $2048 \times 6$ \\
    Number of epochs  & $5$ \\
    Learning rate & $1\times 10^{-3}$  \\
    \bottomrule[1.1pt]
    \end{tabular}
    \vspace{-15px}
\end{table}

\begin{figure}[b]
	\centering
	\subfigure[slopes]{
		\begin{minipage}[b]{0.45\linewidth}
			\includegraphics[width=\linewidth]{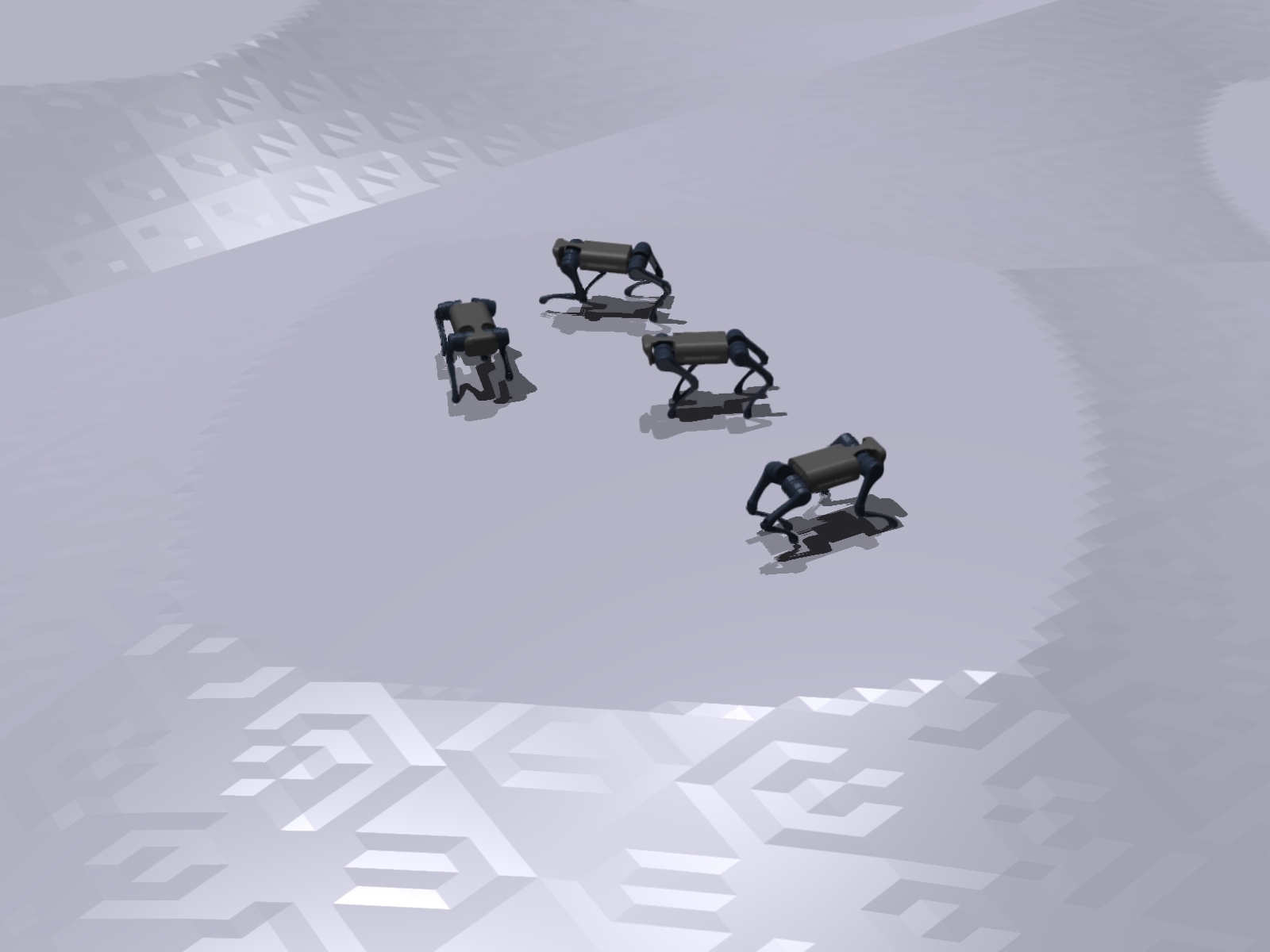}
		\end{minipage}
	}
	\subfigure[rough slopes]{
		\begin{minipage}[b]{0.45\linewidth}
			\includegraphics[width=\linewidth]{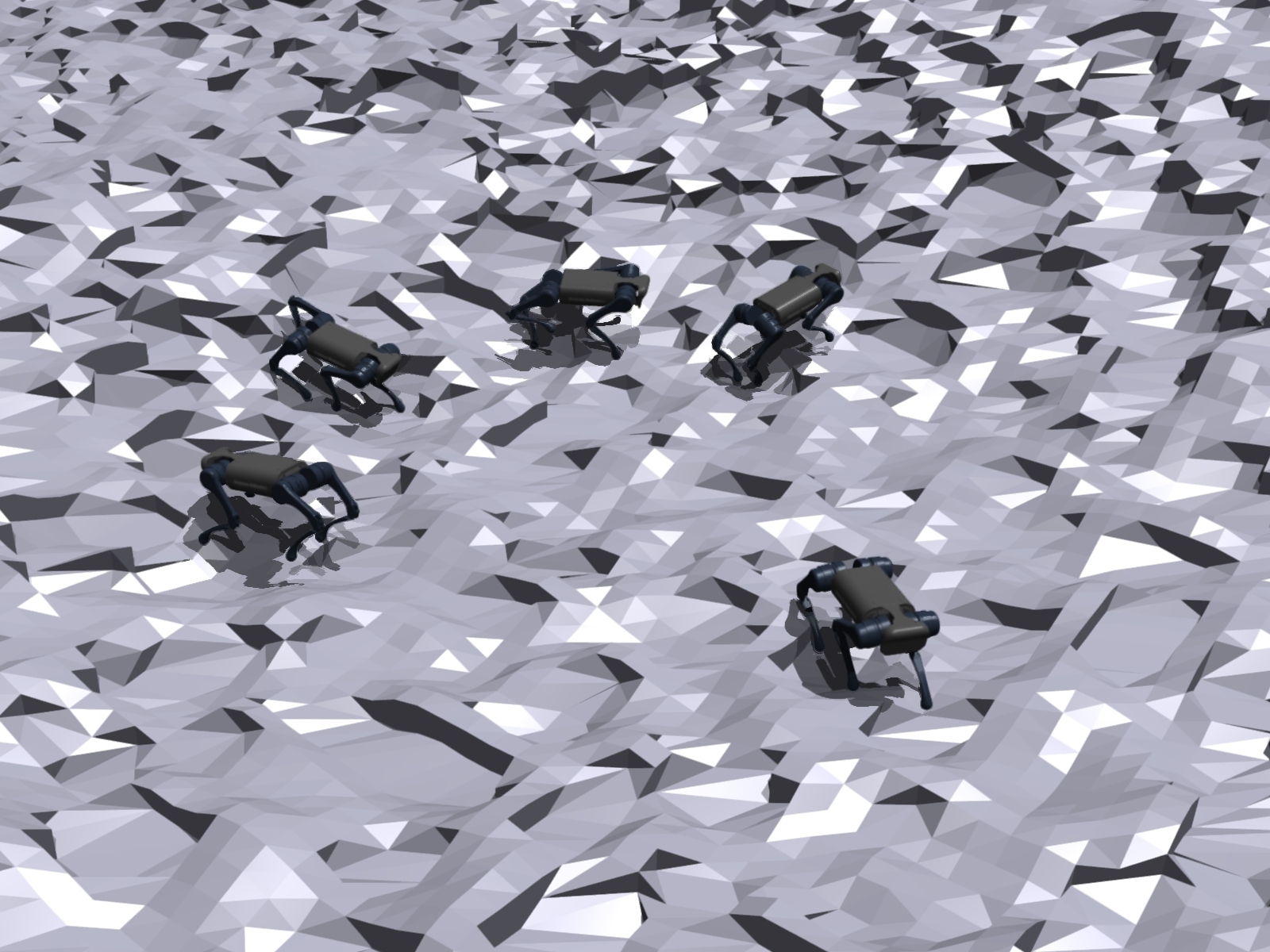}
		\end{minipage}
	}
	\subfigure[stairs]{
		\begin{minipage}[b]{0.45\linewidth}
			\includegraphics[width=\linewidth]{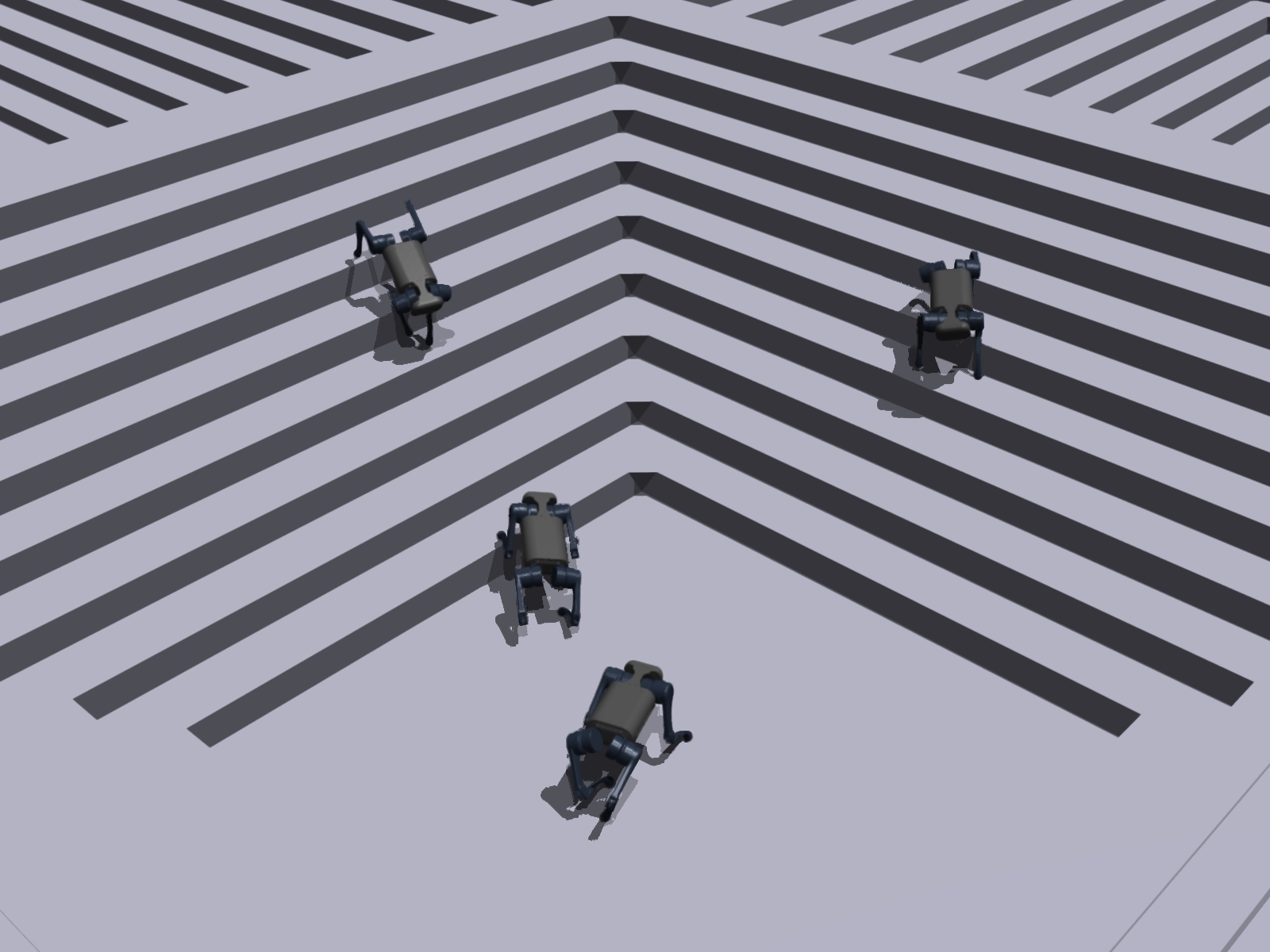}
		\end{minipage}
	}
	\subfigure[discrete obstacles]{
		\begin{minipage}[b]{0.45\linewidth}
			\includegraphics[width=\linewidth]{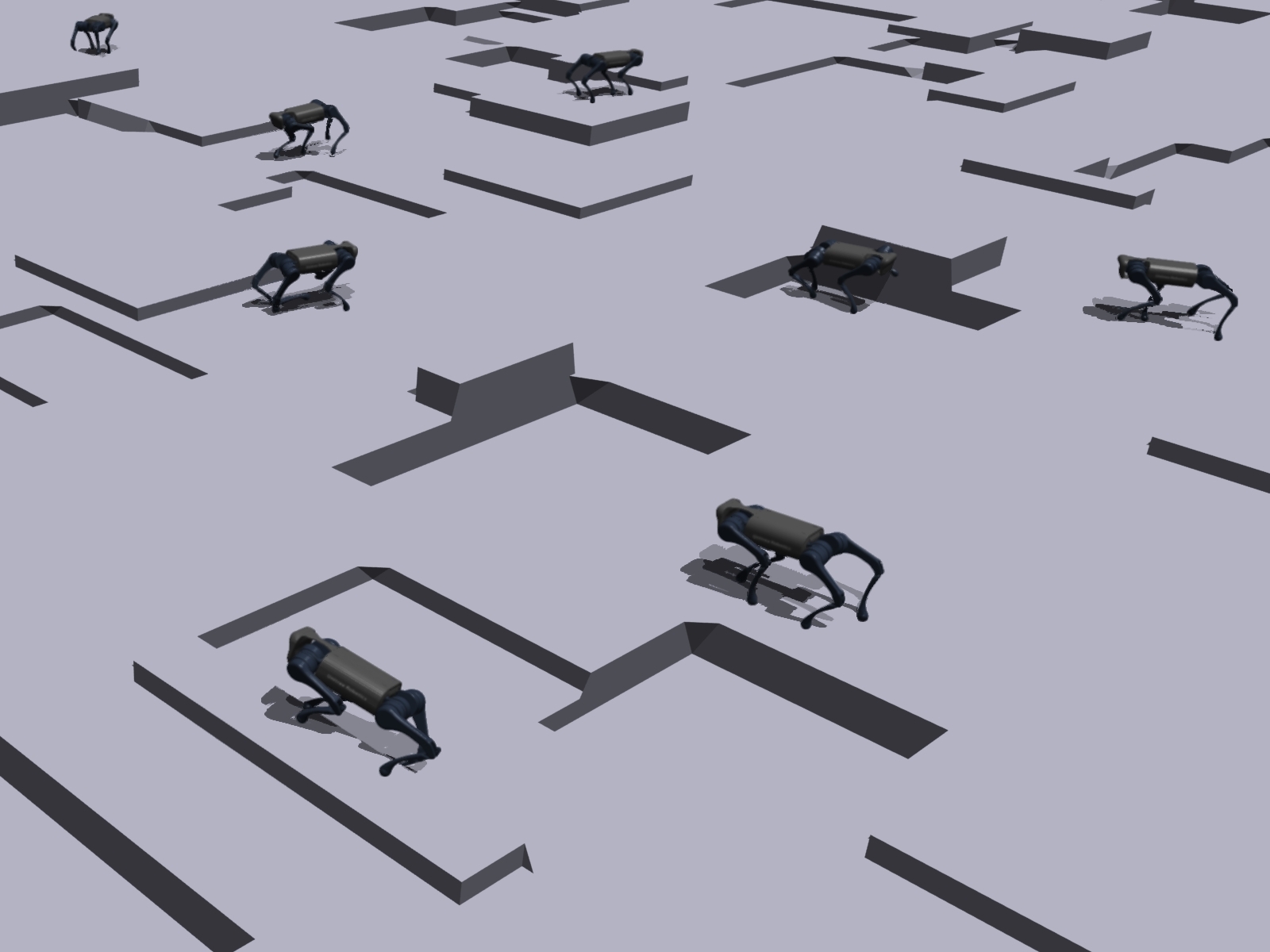}
		\end{minipage}
	}
	\caption{Terrains in simulation.}
	\label{fig:sim_terrain}
\end{figure}

To achieve robust locomotion in various terrain, it is crucial to implement a proper training curriculum strategy. We adopt a terrain curriculum similar to~\cite{rudin2022learn_in_minutes} and have selected four terrain types for our training: slopes, rough slopes, stairs, and discrete obstacles, as shown in Fig.~\ref{fig:sim_terrain}.  The slopes and rough slopes have gradients ranging from \(0^\circ\) to \(26.57^\circ\), with rough slopes additionally containing uniform noise ranging from 5 cm to 17 cm. The stairs have heights ranging from 5 cm to 23 cm, and the discrete obstacles have heights ranging from 5 cm to 24 cm. Each terrain type is divided into difficulty levels from 0 to 9, evenly distributed within the specified difficulty ranges. At the beginning of training, all robots are assigned the lowest difficulty level of these four types of terrain. The robots are moved to more difficult terrains once they have traversed the current area.

Velocity commands are uniformly and randomly sampled from a range $[-1, 1]$ m/s at the beginning. Once they step out of the most difficult terrain and performed velocity tracking well, the velocity commands sampling range is incrementally increased to foster more agile movement skills.

To compensate for the gap between simulation and the real world, we randomize inertial parameters such as the mass of the base and legs, the CoM of the base, the friction and restitution between the rigid body and the ground, the PD gains and motor strength, and action delay. The details of the randomization are presented in Table~\ref{tab:domin rand}.

\begin{table}[t]
    \centering
    \caption{Domain Randomization}
    \label{tab:domin rand}
    \begin{tabular}{lll}
    \toprule[1.1pt]
    Randomization Term & Range &  Unit \\ 
    \hline 
    Link mass  & $[0.8, 1.2]\times \text{nominal value}$ & Kg \\
    Payload mass  & $[-1, 3]$ & Kg \\
    CoM of base  & $[-7.5,7.5]\times[-5,5]\times[-5,5]$ & cm \\
    Friction & $[0.2, 1.7]$ & - \\
    Restitution & $[0.25, 0.75]$ & - \\
    Joint $K_p$  & $[0.8, 1.2]\times \text{nominal value}$ & N$\cdot$rad \\
    Joint $K_d$  & $[0.8, 1.2]\times \text{nominal value}$ & N$\cdot$rad/s \\
    Motor Strength $K_d$  & $[0.8, 1.2]\times \text{nominal value}$ & N \\
    Action delay & $[0, 20]$ & ms \\
    \bottomrule[1.1pt]
    \end{tabular}
    \vspace{-15px}
\end{table}
\begin{figure}[t]
    \centering
    \includegraphics[width=0.8\linewidth]{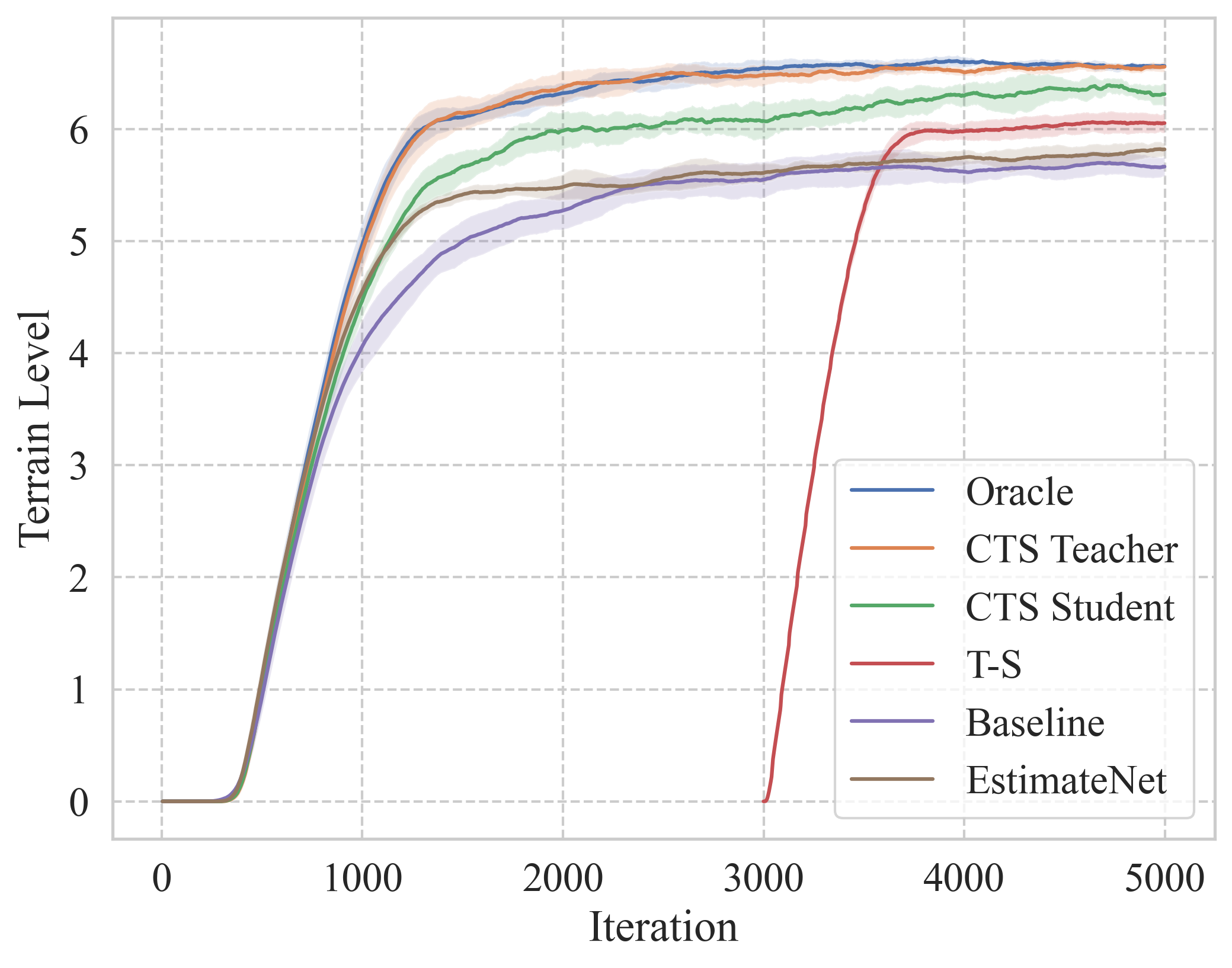} 
    \caption{\footnotesize Learning curves of average terrain level.}
    \label{fig:terrain_level}
    \vspace{-20px}
\end{figure}

\section{Results}\label{sec:results}

\subsection{Evaluation Results}\label{sec:evaluation}
For a comparative evaluation, we compared the training results of these algorithms for the A1 robot as follows:
\begin{itemize}
    \item Oracle: Policy receives the encoded privileged state as input and was trained by PPO.
    \item Baseline: Policy with proprioceptive encoder was trained by PPO.
    \item EstimatorNet: Policy was concurrently trained with an explicit estimator network estimating body velocity and feet height similar to~\cite{Hwangbo2022vel_est}.
    \item Two-stage teacher-student (T-S): The proposed method trained in two stages where student policy was trained by supervised learning using latent reconstruction loss and action imitation loss, following the original teacher-student learning framework as described in~\cite{joonholee2020scirobotics}.
    \item Regularized Online Adaptation (ROA): A single-stage training method presented in~\cite{roa}.
\end{itemize}
For a fair comparison, all the methods above were trained within asymmetric actor-critic framework under the same training configuration detailed in Section~\ref{sec:training setup}, and use the same network scale and random seeds. Our evaluations will employ the policy at the 5000th iteration. Specifically, for the two-stage teacher-student method, the policy is obtained by 3000 iterations for teacher and 2000 iterations for student. For ROA, we obtain the same switch period and regularization curriculum as described in~\cite{roa}.
\begin{figure}[b]
    \centering
    \includegraphics[width=0.9\linewidth]{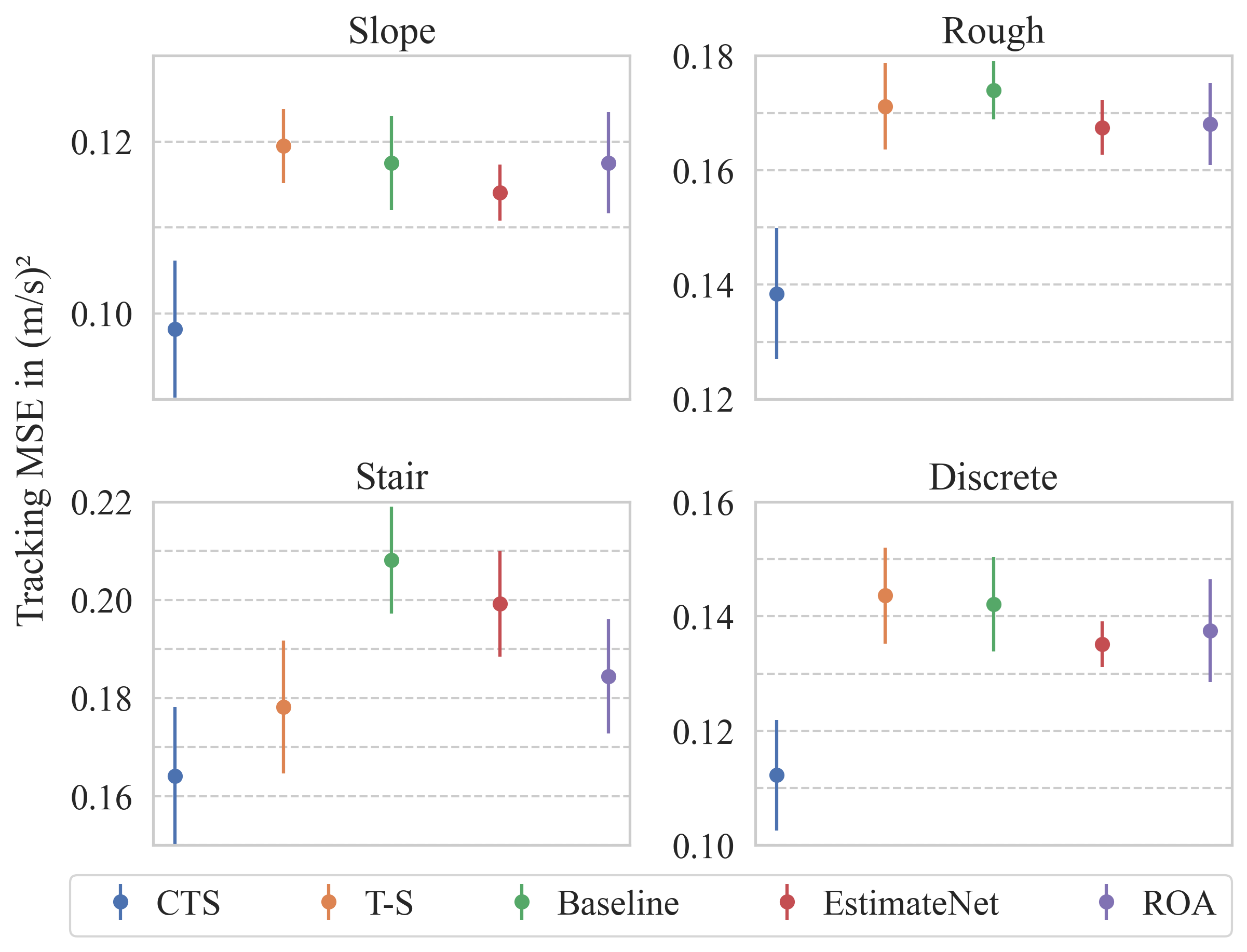} 
    \caption{\footnotesize Evaluation of average tracking error in four types of terrains. Linear velocity commands were uniformly sampled from $[-1.0, 1.0]$ m/s.}
    \label{fig:tracking_mse}
\end{figure}

\begin{figure}[t]
    \centering
    \includegraphics[width=0.9\linewidth]{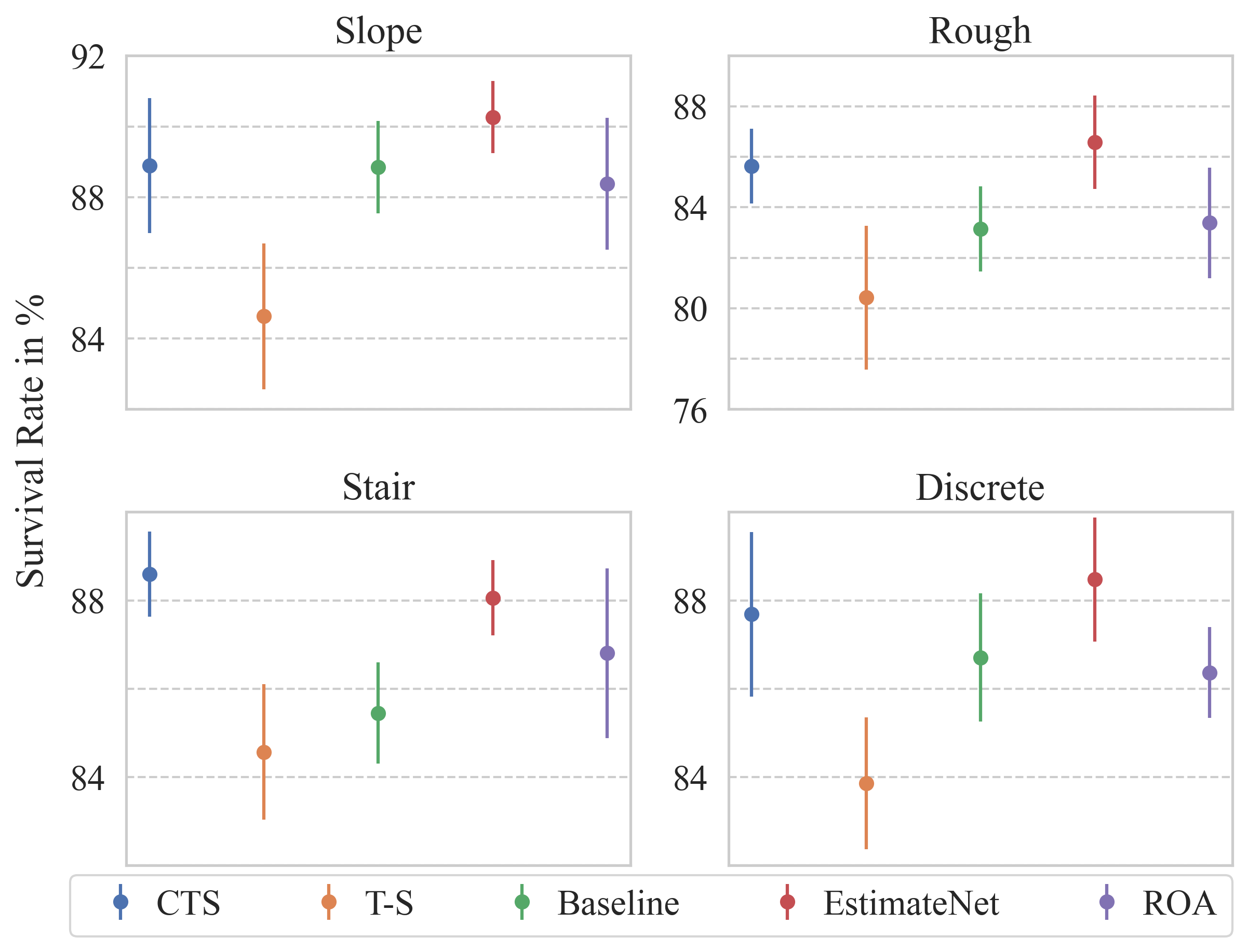} 
    \caption{\footnotesize Evaluation of push recover in four types of terrains. Push by applying force inducing a velocity change of approximately 2.5m/s of the robots.}
    \label{fig:push_recover}
    \vspace{-10px}
\end{figure}
We compared improvements in terrain level during the training process. The results are shown in Fig.~\ref{fig:terrain_level}, in which the curves are averaged over 5 seeds. The shaded area represents the standard deviation across seeds. The terrain level curve records the average terrain level of all agents at each moment during the training process. It serves as a statistical measure of the agents' overall mobility capabilities. Fig.~\ref{fig:terrain_level} shows that teacher in CTS exhibits performance nearly identical to the Oracle, indicating that synchronous training with the student does not impair the teacher's performance. Student in CTS performs slightly worse than the teacher but still outperforms the student policy trained in two-stage teacher-student by imitating teacher. The baseline, which does not directly utilize privileged information in agent decision-making, exhibits similar final performance to EstimateNet. However, due to the guidance provided by supervised learning signals for body velocity and feet height, EstimateNet shows faster initial learning compared to the baseline. Due to the policy-switching characteristic of ROA during training, the terrain level curve does not effectively reflect the mobility capabilities of individual policies and lacks reference value; thus, it has been excluded from consideration.
\begin{figure}[t]
    \centering\includegraphics[width=0.9\linewidth]{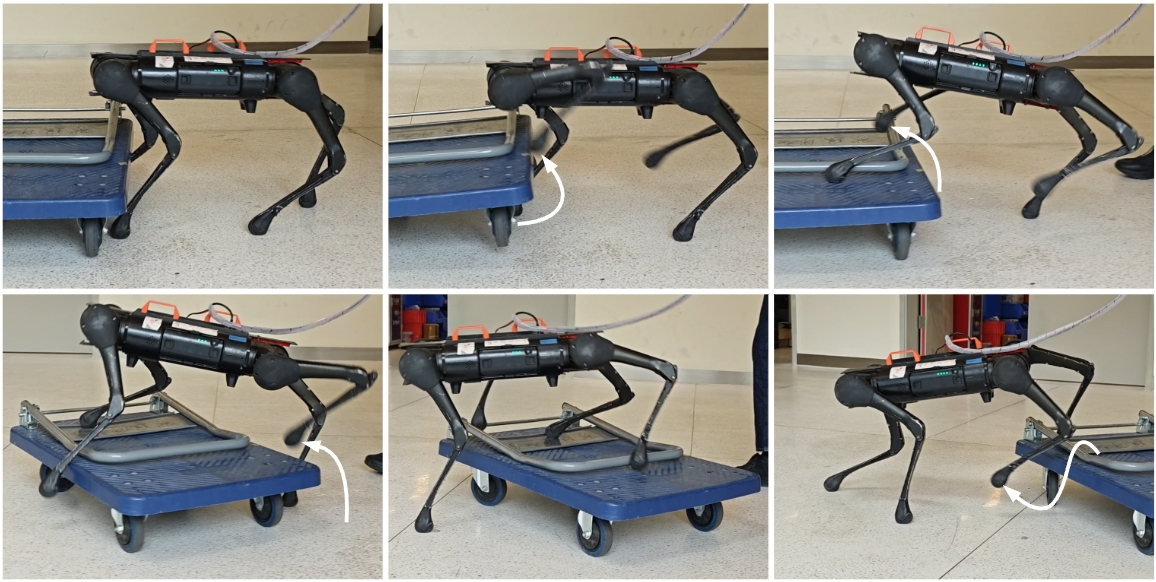} 
    \caption{\footnotesize Aliengo steps over a moving platform.}
    \label{fig:aliengo_slip}
    \vspace{-15px}
\end{figure}

We evaluated velocity tracking performance under various terrain conditions by distributing 8192 robots evenly among four types of terrains. Velocity commands were uniformly sampled from $[-1.0, 1.0]$ m/s. Tracking errors are quantified using the metric $\Vert\boldsymbol{v}_{xy}^\text{cmd} - \boldsymbol{v}_{xy}\Vert_2$. Fig.~\ref{fig:tracking_mse} presents the average tracking error (y-axis) of different training methods under four types of terrains. The points represent the average values of the policies trained under 5 different seeds, while the lines indicate the standard deviation. It shows that student of CTS achieves a reduction in velocity tracking error compared to the student trained in two-stage teacher-student, with improvements of 17.85\% on slopes, 19.12\% on rough slopes, 7.9\% on stairs, and 21.85\% on discrete obstacles. On the stair terrain, the velocity tracking performance of the baseline and EstimateNet is relatively poor compared to other methods. We believe this is because the stair terrain is the most challenging for maintaining good velocity tracking among all terrains. Therefore, it requires the terrain information from the privileged information more significantly, which explains why the three methods directly guided by privileged information exhibit better velocity tracking performance on the stair terrain.


\begin{figure*}[h]
    \centering
    \includegraphics[width=0.8\linewidth]{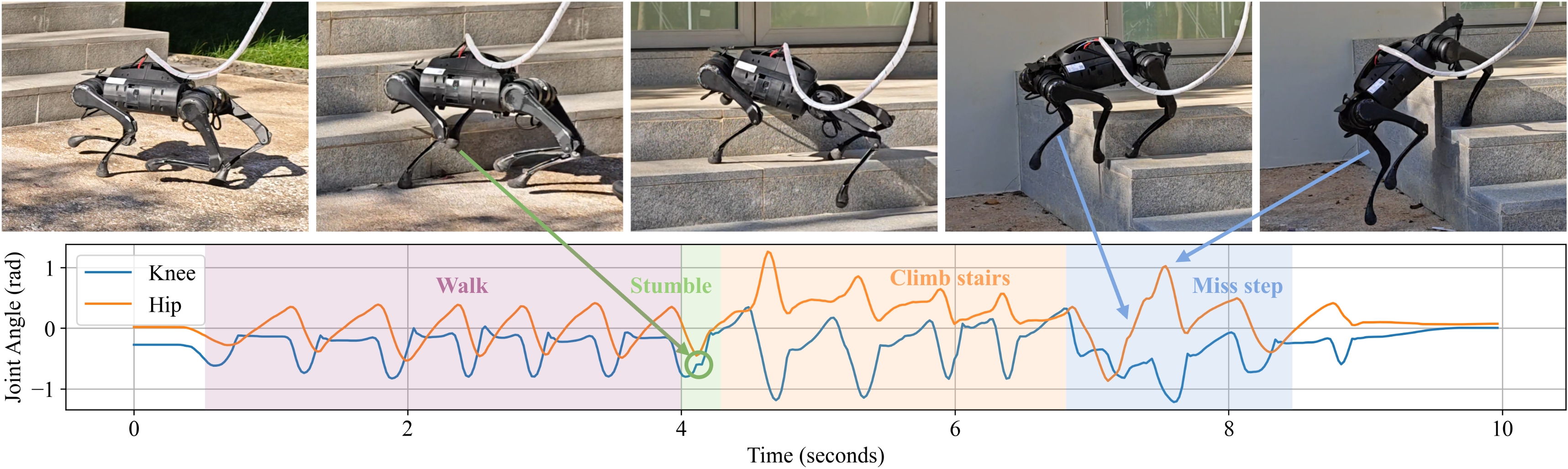} 
    \caption{\footnotesize Quadruped locomotion with stairs and missing step adaptation. The curves record the joint angles of the robot's right front leg.}
    \label{fig:a1_stair_test}
    \vspace{-10px}
\end{figure*}
\begin{figure*}[h]
    \centering
    \includegraphics[width=0.8\linewidth]{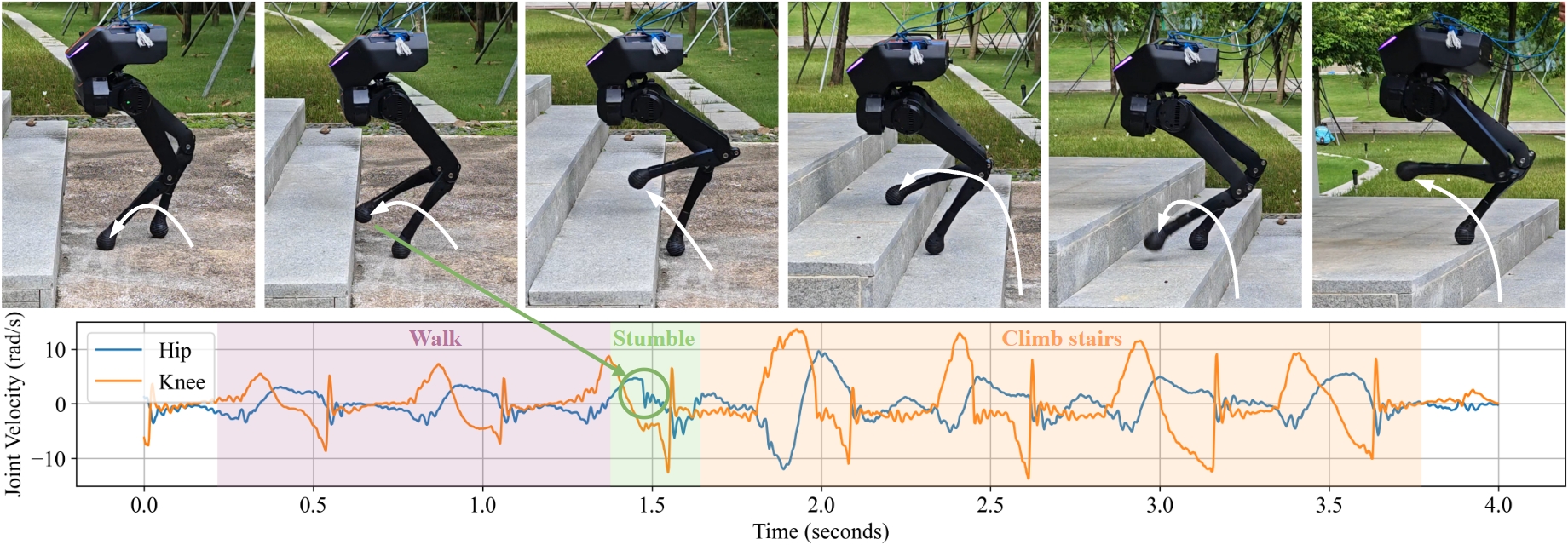} 
    \caption{\footnotesize Biped locomotion with stairs. The curves record the joint angles of the robot's right leg.}
    \label{fig:pf_stair_test}
    \vspace{-15px}
\end{figure*}

To evaluate the robustness of policies, we apply random force to the robot’s body and record their survival rates on various terrains. Specifically, the direction of the push was chosen randomly, with the applied force inducing a velocity change of approximately 2.5m/s of the robots. Results in Fig.~\ref{fig:push_recover} shows CTS demonstrates greater robustness across various terrains compared to the two-stage teacher-student. The survival rates under random push for the four types of terrain are higher by 5.04\%, 6.47\%, 4.76\%, and 4.57\%, respectively. The staged training student exhibits the poorest robustness against random pushes. This may be due to its sole training objective of imitating the teacher, without considering the reinforcement learning goal of maximizing expected discounted return. Due to the information gap, the student cannot perfectly mimic the teacher, ultimately leading to a lack of emphasis on maintaining balance and making it more prone to being toppled by pushes. The figure also shows that EstimateNet performs quite well across various terrains, which may benefit from its explicit estimation of the base linear velocity. We believe that adding a similar mechanism to our method will help further improve robustness.


Based on the results of the comparative evaluation, we believe that in training policies that can only access proprioceptive observations, directly incorporating privileged information to aid decision-making, along with reinforcement learning aimed at maximizing the expected discounted return, helps in achieving robust and high-performance policies.

\subsection{Real-World Experiments}\label{sec:realworld}
We implemented the student policy on quadruped robots of varying sizes, including the Unitree A1 and Aliengo, as well as on the more challenging point-foot bipedal robot, LimX Dynamics P1. All demonstrated exceptional robustness and terrain traversal capabilities, validating the universality and superior performance of our method.

Fig.~\ref{fig:aliengo_slip} shows the quadruped robot is engaged in a complex interaction with a moving platform, providing a clear illustration of its robustness against uncertainty and terrain adaptability. Initially, the robot maintains regular walking towards the platform. When the robot's left front leg encounters an obstacle, it instinctively retracts and lifts to step onto the platform, a response not explicitly trained for since the stairs used during training did not feature gaps. This behavior demonstrates the robot's generalization capabilities, allowing it to adapt to new environmental challenges despite the absence of direct prior experience with such specific scenarios. Once the left front leg secures placement on the platform, the robot is able to discern the platform's elevation, thereby enabling the right front leg to accurately follow suit and step onto the platform without undesirable collision. 

As the platform commences sliding upon the robot's ascent, the robot's hind legs promptly engage, adjusting to maintain balance in response to the new dynamics introduced by the moving platform. This demonstrates the robot's reactive balance capabilities and its proficiency in stabilizing itself amid changing environmental conditions. As the robot descends from the platform, the placement of its left hind leg is dynamically adjusted in relation to the robot's overall posture and the positioning of its other legs. During its swing phase, it selects a more forward foothold to align with the robot's overarching motion, resulting in an S-shaped swing trajectory. 

Fig.~\ref{fig:a1_stair_test} provides visual and data analysis illustrating the quadruped robot’s motion as it encounters stairs and missing steps. The curves in the figure represent the angular trajectories of the hip and knee joints of the robot's right front leg. When the robot encounters a step while walking, its right front leg is quickly lifted after being stumbled, allowing it to step onto the stair. The robot then smoothly reaches the top and proceeds toward the edge until its front leg steps into empty space, initiating a fall. Immediately sensing the change in terrain, the robot reacts swiftly, extending its front legs to seek additional support and keeping its center of mass within a safe zone determined by the stance of its limbs. Once the front legs make contact, the hind legs follow, enabling the robot to safely return to the ground.

Fig.~\ref{fig:pf_stair_test} demonstrates the effectiveness of our policy applied to a more challenging bipedal platform. The curves in the figure represent the angular velocity trajectories of the hip and knee joints of the robot's right leg. The policy enables the robot to maintain stable forward locomotion on flat surfaces. When the right leg hits the edge of a step during the swing phase (as indicated by the instantaneous decrease in the angular velocity of the hip joint in the figure), the swing trajectory is forced to change, causing the foot to land prematurely. At this point, the policy has already detected the presence of the obstacle through proprioceptive observation. Subsequently, the left leg is raised directly to overcome the obstruction. After the left leg lands, the policy perceives the approximate height of the stair, allowing subsequent swing trajectories to have sufficient height to smoothly ascend the stair.

\section{Conclusions and Future Works}\label{sec:conclusion}
In this study, we present the Concurrent Teacher-Student Learning framework, designed to equip legged robots with the capability to navigate unstructured terrains using purely proprioception. The effectiveness of this framework has been validated through its implementation on differently sized quadruped robots, showcasing their ability to maneuver over moving platforms and stairs. Even for bipedal robots with completely different configurations and higher degrees of underactuation, the strategies trained through our method also achieve excellent robustness and terrain traversal capabilities. A notable limitation of this approach is the requirement for physical leg-obstacle interaction for adaptation. Future efforts will focus on incorporating exteroceptive inputs into the locomotion system, aiming to refine gait planning and obstacle negotiation strategies before physical contact occurs.

\bibliographystyle{IEEEtran}
\bibliography{Bibliography}

\end{document}